\documentclass{article}

\usepackage{arxiv}
\usepackage[utf8]{inputenc}
\usepackage[T1]{fontenc}
\usepackage{hyperref}
\usepackage{url}
\usepackage{booktabs}
\usepackage{amsfonts}
\usepackage{amsmath,amssymb}
\usepackage{microtype}
\usepackage{graphicx}
\usepackage{dcolumn}
\usepackage{bm}
\usepackage{natbib}
\usepackage{doi}
\usepackage{CJKutf8}

\title{Turbulence-like $5/3$ spectral scaling in contextual representations of language as a complex system}

\author{
Zhongxin Yang \\
College of Engineering, Peking University, Beijing, 100871, China
\And
Chun Bao \\
College of Engineering, Peking University, Beijing, 100871, China
\And
Yuanwei Bin\thanks{Corresponding author: \texttt{ybin@eitech.edu.cn}} \\
Ningbo Institute of Digital Twin, Eastern Institute of Technology, Ningbo, 315200, Zhejiang, China \\
Shenzhen Tenfong Technology Co., Ltd., Shenzhen 518000, China
\And
Xiang I. A. Yang\thanks{Corresponding author: \texttt{xzy48@psu.edu.cn}} \\
Mechanical Engineering, Pennsylvania State University, State College, PA, 16802, USA
\And
Shiyi Chen \\
College of Engineering, Peking University, Beijing, 100871, China \\
Ningbo Institute of Digital Twin, Eastern Institute of Technology, Ningbo, 315200, Zhejiang, China
}

\date{\today}

\begin{document}
\maketitle

\begin{abstract}
Natural language is a complex system that exhibits robust statistical regularities. Here, we represent text as a trajectory in a high-dimensional embedding space generated by transformer-based language models, and quantify scale-dependent fluctuations along the token sequence using an embedding-step signal. Across multiple languages and corpora, the resulting power spectrum exhibits a robust power law with an exponent close to $5/3$ over an extended frequency range. This scaling is observed consistently in contextual embeddings from both human-written and AI-generated text, but is absent in static word embeddings and is disrupted by randomization of token order. These results show that the observed scaling reflects multiscale, context-dependent organization rather than lexical statistics alone. By analogy with the Kolmogorov spectrum in turbulence, our findings suggest that semantic information is integrated in a scale-free, self-similar manner across linguistic scales, and provide a quantitative, model-agnostic benchmark for studying complex structure in language representations.
\end{abstract}


%\tableofcontents

%\boxedtext{
% Research reports require a significance statement of between 50 and 120 words. Abbreviations are permitted, but citations cannot be included. If required, un-comment this element in the template to include. The heading is included automatically.
%}

\section{Introduction}
% \textit{Introduction.}
Natural language is a highly complex system. Its hierarchical structure, which balances strict grammatical constraints with immense expressive variability, manifests in robust statistical regularities across languages and domains.
Classic empirical laws, such as Zipf's law for heavy-tailed word frequencies \cite{zipf2016human}, provide macroscopic constraints on linguistic organization.
Furthermore, recent work on linguistic ``criticality'' suggests that text may balance order and randomness in a regime that supports rich variability \cite{lin2017critical,nakaishi2024critical}.
Long-range correlations in texts have also been linked to hierarchical linguistic and semantic organization \cite{altmann2012origin}.
Together, these findings motivate the view of language as a complex system with structure spanning multiple scales.
However, most existing laws are formulated in terms of observable text statistics, and it remains unclear whether \emph{representation-level} signatures---especially those learned by modern neural language models---exhibit similarly simple and reproducible multiscale behavior.

Transformer-based language models \cite{vaswani2017attention} provide a natural setting to address this question.
Models such as BERT \cite{devlin2019bert} and GPT-style transformers \cite{brown2020gpt3} map a token sequence into a sequence of high-dimensional vectors (contextual token representations) whose values depend on surrounding words.
This contextual dependence distinguishes transformers from static word embeddings such as Word2Vec \cite{mikolov2013distributed} and GloVe \cite{pennington2014glove}.
Contextual embeddings thus offer a model-defined representation of language-in-context, enabling quantitative analysis beyond token counts alone.

Spectral analysis provides a compact description of multiscale organization in complex systems, and power laws can serve as empirical fingerprints.
An exponent near five-thirds is well known from the Kolmogorov spectrum in turbulence \cite{kolmogorov1991local}, characterizing the scale-free cascade of energy.
Similar scaling has been reported in atmospheric spectra \cite{nastrom1985climatology},
%\cite{nastrom1985climatology,tulloch2006theory}, 
solar-wind turbulence \cite{goldstein1995mhd}, 
%\cite{goldstein1995mhd,roberts2010evolution}, 
and interstellar electron-density fluctuations \cite{armstrong1995electron}. %.
%Not assuming shared mechanisms across these systems, 
Here, we also use spectral scaling as a model-agnostic tool to characterize multiscale structure in contextual language representations.

In this letter, we analyze texts as sequences of contextual token representations and quantify scale-dependent structure along token position.
To emphasize local variation while reducing sensitivity to slow document-level drift, we study token-to-token embedding differences (the embedding-step signal) rather than the raw embedding trajectory.
Across multiple languages and diverse corpora---including both human-written and AI-generated text---the power spectrum of embedding-step changes follows a robust power law with an exponent close to 5/3 over an extended frequency range.
This scaling is consistently observed for transformer-based contextual embeddings, but is absent in static embeddings and is eliminated by randomizing token-representation order, indicating that it arises from context-dependent organization rather than lexical statistics alone.
These results identify a reproducible spectral fingerprint of contextual language embeddings and provide a quantitative benchmark for comparing multiscale structure across languages, corpora, and model families.

\section{Method}
% \textit{Method.}
We constructed corpora spanning four languages (Chinese, English, German, and Japanese) and two sources (human-authored and AI-generated).
To reduce sensitivity to document length while ensuring consistent spectral resolution, all samples were truncated to a fixed length of $L=1200$ tokens.
Human-authored corpora cover multiple domains. AI-generated corpora were produced using the \texttt{DeepSeek-V3} model. 
%For Chinese, we used scientific writing collected from CNKI (primarily 2024--2025), news from CCTV (2022--2025), and a movie-review corpus from Douban (The Shawshank Redemption) \cite{cnki,cctv,douban}. 
%For English, human scientific texts were collected from scientific journals including Physical Review Letters, arXiv, Journal of Fluid Mechanics, and Physics of Fluids, while human news and reviews were drawn from cleaned CNN articles (2011--2022) and the IMDB 50k movie-review dataset, respectively \cite{cnn_dataset,imdb_dataset}.
%For Japanese, human news samples were taken from a cleaned CC-News dataset (2024) \cite{ccnews_ja}.
%For German, human news samples were taken from the 10kGNAD dataset (2015--2016) \cite{gnad_de}.
%AI-generated corpora were produced using the \texttt{DeepSeek-V3} model.
%For each language, prompts were designed to generate long-form text in the same broad categories as the human corpora, including scientific-style writing and news reports; in Chinese we additionally generated Gaokao-style essays.
%Prompts enforced minimum length requirements (typically $\ge 800$ Chinese characters or $\ge 600$--$800$ English/German/Japanese words) and specified genre-specific structure (e.g., headline/lead/body for news) without post-editing \cite{deepseekai2024deepseekv3technicalreport}.
Representative topic lists and prompt templates are provided in the Appendix \ref{app}.

%\subsection*{Embedding models and representation extraction}
We embedded each token sequence using both contextual and static embedding models.
Contextual representations were obtained from transformer-based models, including language-specific BERT variants for Chinese, German, and Japanese, and an open-source GPT-style model \cite{openai2025gptoss120bgptoss20bmodel} for English.
Static embeddings were obtained from Word2Vec (Chinese) and GloVe (English).
All models were accessed through Hugging Face or official repositories \cite{gloveweb,word2vecweb}.

For each text sample, tokenization followed the native tokenizer of the embedding model.
Let $\mathbf{x}(t) \in \mathbb{R}^{d}$ denote the contextual token embedding at token position $t$, where $d$ is the embedding dimension.
Unless otherwise stated, $\mathbf{x}(t)$ was taken from the final hidden layer of the contextual model.  % [TODO: confirm layer choice if not final]

%\subsection*{Spectral analysis of embedding-step signals}
To quantify scale-dependent structure along token position while reducing sensitivity to slow document-level drift, we analyzed the token-to-token embedding difference signal:
\begin{equation}
\mathbf{v}(t) = \mathbf{x}(t+1) - \mathbf{x}(t),
\end{equation}
which we interpret as informational fluctuation of the semantic trajectory.
For each sample and each embedding dimension, we computed the discrete Fourier transform of $\mathbf{v}(t)$ along $t$, yielding $\mathbf{u}(f)$.
The one-dimensional power spectral density was then formed by taking the ensemble average across the $d$ spatial dimensions and over the corpus samples. To allow universal comparison across different models and dimensionalities, the dimension-averaged spectrum $\overline{E}(f)$ was normalized by the total variance, $\sigma^2 = \int \overline{E}(f)\, \mathrm{d}(f/f_{\max})$.

\begin{figure}[htb!]
    \centering
    \includegraphics[width=0.7\linewidth]{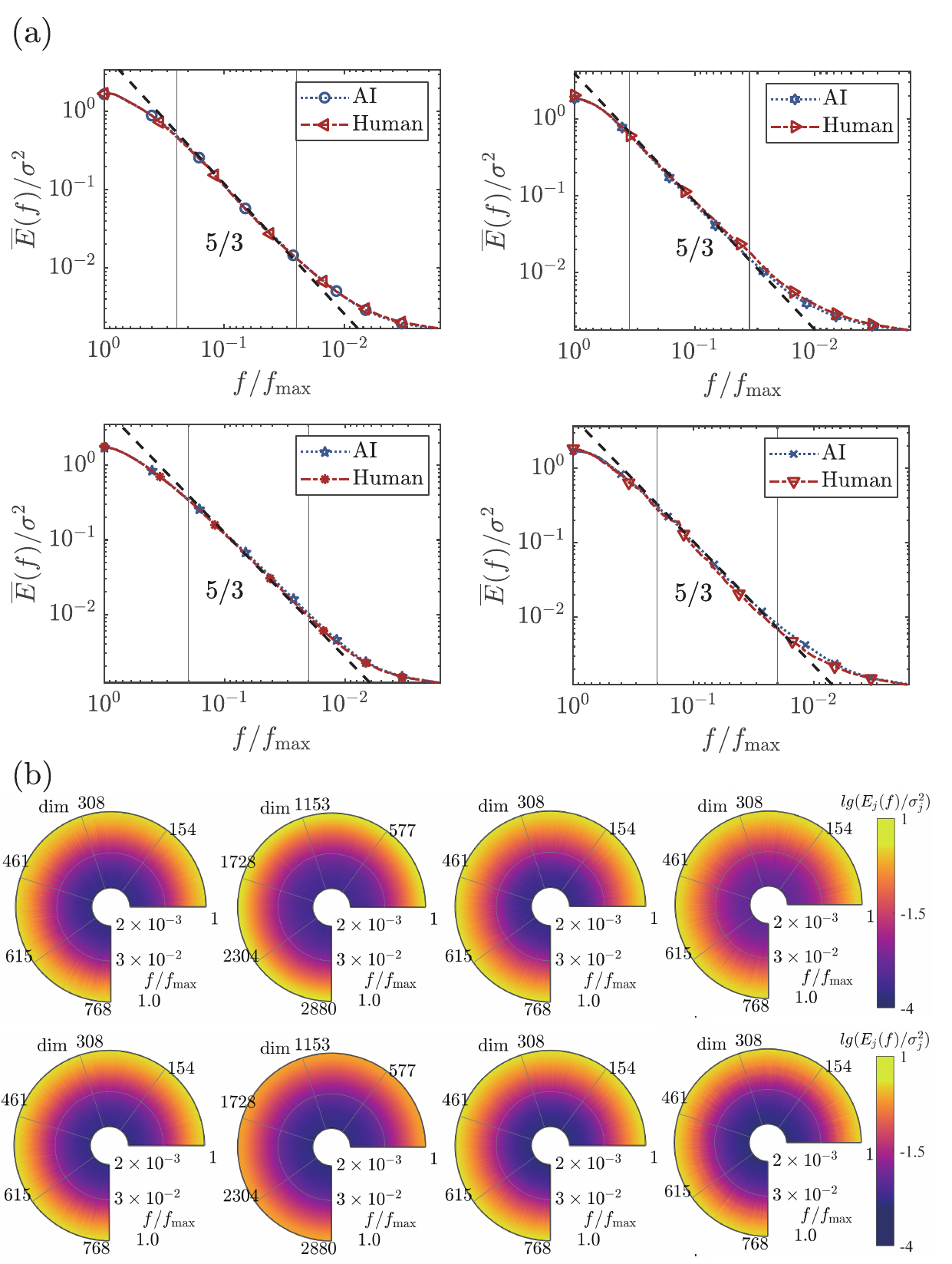}
  
    \caption{(a) {Power spectra of contextual embedding-step sequences.} The spectra are normalized by the variance $\sigma^2 = \int_{0}^{1} \overline{E}(f)\, \mathrm{d}(f/f_{\max})$. The dashed line indicates the $+5/3$ slope.
    The four panels are Chinese/BERT (top left); English/GPT-OSS (top right); German/BERT (bottom left); Japanese/BERT (bottom right).
    %\textit{(a)} Chinese/BERT; \textit{(b)} English/GPT-OSS; \textit{(c)} German/BERT; \textit{(d)} Japanese/BERT.
    (b) Polar heatmaps of normalized per-dimension power spectra. The radial axis represents frequency (log scale), and the angular axis represents embedding dimension index.
    Top row: AI, bottom row: human; From right to the left: Chinese, English, German, Japanese.
    %\textit{(a)} Chinese/BERT/AI; \textit{(b)} Chinese/BERT/Human;
    %\textit{(c)} English/GPT-OSS/AI; \textit{(d)} English/GPT-OSS/Human;
    %\textit{(e)} German/BERT/AI; \textit{(f)} German/BERT/Human;
    %\textit{(g)} Japanese/BERT/AI; \textit{(h)} Japanese/BERT/Human.
    }
    \label{fig:spectrum}
\end{figure}

\section{Results}
% \textit{Results.}
Figure~\ref{fig:spectrum}a shows the dimension-averaged spectrum
\begin{equation}
\overline{E}(f)=\frac{1}{d}\sum_{j=1}^{d}E_j(f),
\end{equation}
for contextual embedding in four language/model configurations.
For each document, we compute $\overline{E}(f)$ and then average the resulting spectra across all documents within the same configuration.
Across all configurations, the spectra exhibit a clear power-law regime over an extended frequency range, and the human and AI curves closely collapse in each panel (Fig.~\ref{fig:spectrum}a).
Here, we estimate the scaling exponent by an ordinary least-squares fit on log--log axes over the frequency window indicated by the two vertical lines (approximately $f/f_{\max}\in[2\times10^{-1},\,2\times10^{-2}]$), which spans about one decade.
Within this window, the fitted exponent is consistently close to five-thirds.

The collapse of the averaged spectra raises a natural question: is the observed scaling confined to a small subset of embedding dimensions, or is it broadly expressed across the representation space?
To assess this, we compute the per-dimension spectrum $E_j(f)$ for each component $j=1,\ldots,d$ and visualize the normalized spectra as polar heatmaps (Fig.~\ref{fig:spectrum}b), with frequency on the radial axis and embedding-dimension index on the angular axis.
Across languages and architectures, the heatmaps reveal similar spectral structure over a large fraction of dimensions, indicating that the scaling is not driven by a small number of dominant axes.
Consistent with this visual evidence, more than 90\% of dimensions yield fitted exponents within $\pm 10\%$ of $5/3$.

\begin{figure}[h]
    \centering
    \includegraphics[width=0.8\linewidth]{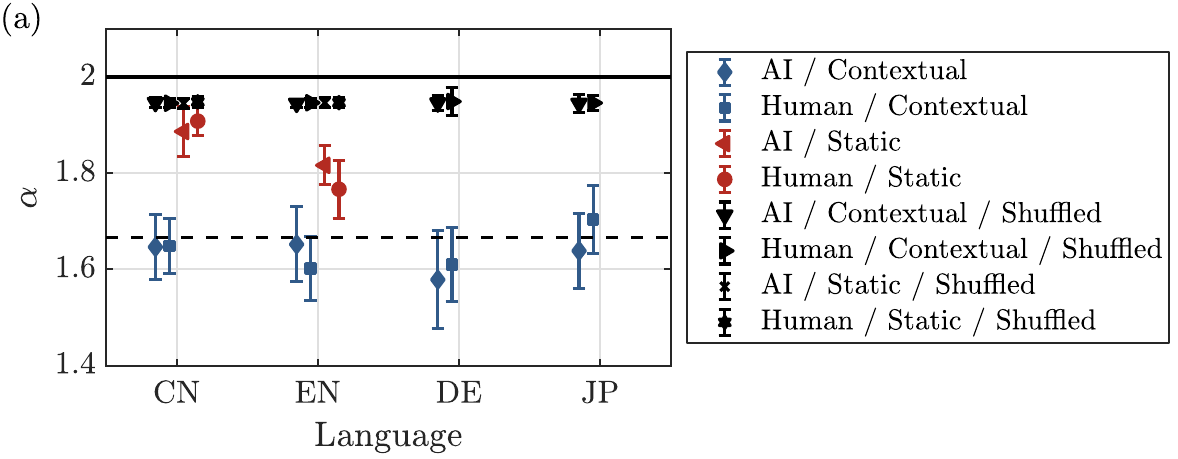}
    \includegraphics[width=0.64\linewidth]{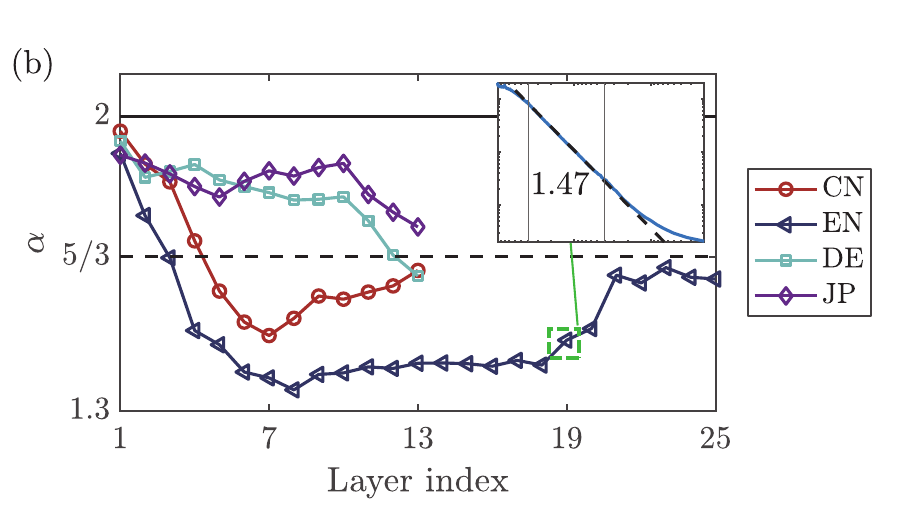}
    \caption{   
    (a) Summary of fitted scaling exponents $\alpha$. `CN', `EN', `DE', and `JP' denote Chinese, English, German, and Japanese, respectively. Error bars show variability across embedding dimensionalities (one standard deviation).
    (b) Fitted power-law exponent versus transformer layer index for human-authored text.
    %The dashed line marks $\alpha=5/3$.
    %and the gray band denotes a tolerance of $\pm5\%$ around $5/3$.
    %Shaded envelopes indicate variability (one standard deviation) across documents within each configuration.
    The inset shows a representative mid-layer spectrum (highlighted by the green box) with a fitted power-law exponent $\alpha\approx 1.47$.
    }
    \label{fig:summary}
\end{figure}

We next test whether the scaling is specific to contextual processing and ordered token-representation sequences using two contrastive analyses.
First, we replace contextual embeddings with static word embeddings (Word2Vec and GloVe), which do not incorporate surrounding context.
Second, we destroy contextual structure by randomly shuffling the token-representation order within each sequence prior to spectral analysis.
Figure~\ref{fig:summary}a summarizes the fitted scaling exponents across all configurations.
Contextual embeddings cluster near five-thirds for all four languages and for both human-written and AI-generated text.
In contrast, static embeddings do not exhibit the same scaling behavior, and token-representation-order randomization eliminates the five-thirds regime, producing substantially different exponents.
%(For spectrums, please refer to Fig. S1 and Fig. S2).
Together, these controls indicate that the observed scaling is not a generic property of embedding vectors alone, but instead reflects context-dependent organization that depends on sequential order.

Finally, to test whether the five-thirds scaling strengthens as context is progressively incorporated, we repeat the analysis using embeddings extracted from intermediate transformer layers.
Figure~\ref{fig:summary}b plots the fitted exponent versus layer index for human-authored text.
For the BERT configurations (13 layers in the present setup), the exponent moves toward $5/3$ with increasing depth: DE and JP approach $5/3$ in an almost monotonic manner, with a relatively flat plateau near $\alpha\approx 1.85$ around layers 6--9, whereas CN exhibits a mid-layer dip to $\alpha\approx 1.5$ (around layer 7) before recovering toward $5/3$ in later layers.
For the GPT configuration (25 layers), the evolution is similarly non-monotonic: the exponent decreases to $\alpha\approx 1.5$ by roughly layer 7, remains near this level through about layer 19, and then converges toward $5/3$ at layer 21, staying close to $5/3$ through the deepest layers.
This result indicates that the scaling is not a fixed property of token statistics or the embedding space alone, but becomes most consistent in deeper layers where contextual information is most fully integrated.

%\section{Concluding Remarks}
\textit{Concluding Remarks.}
We report a robust five-thirds power-law scaling in the spectrum of token-to-token embedding differences produced by contextual language models.
The scaling is observed across four languages and diverse corpora, including both human-written and AI-generated text, and appears consistently in transformer-based architectures (BERT and GPT-style models).
In contrast, the scaling is absent in static word embeddings (Word2Vec and GloVe) and disappears when token-representation order is randomized, indicating that it depends on contextual representations and sequential structure.
Together, these results identify a reproducible spectral fingerprint of contextual language embeddings.

Power-law spectra provide compact summaries of multiscale organization in complex systems.
A canonical example is the Kolmogorov $-5/3$ spectrum in turbulence, which reflects cross-scale transfer in the inertial range \cite{kolmogorov1991local}.
Motivated by this perspective, our observation of a $+5/3$ scaling suggests that contextual representations exhibit scale-free structure across a wide range of token separations.
Just as the 5/3 spectrum in turbulence reflects a scale-free coupling between large and small eddies, the observed scaling in contextual embeddings suggests that semantic information is integrated in a self-similar manner across linguistic scales—coupling local syntax with broader document context without a single characteristic length scale.
While the present results do not identify an underlying mechanism, they are consistent with the intuition that coherence is assembled from local to progressively larger textual units, and with our finding that the scaling becomes more pronounced in deeper transformer layers.

Beyond its conceptual interest, the scaling has practical implications for model evaluation.
The convergence of different contextual architectures toward a similar exponent suggests that this statistic captures a shared property of contextual representations.
As such, the five-thirds spectrum provides a simple, task-independent diagnostic that can complement downstream benchmarks and enable systematic comparisons across corpora, languages, and model families.

%\section*{Data availability.}~
\textit{Data availability.}
All the datasets used in this study are available in the Zenodo repository at https://doi.org/10.5281/zenodo.18846435, and all the code used in analysis is available in the GitHub repository upon acceptance.

%\section*{Acknowledgements.}~ 
\textit{Acknowledgments.}
All authors declare that they have no conflicts of interest.

\appendix
\section{Corpus Details and Generation Methodology}
\label{app}
We curated a multilingual corpus spanning Chinese, English, German, and Japanese to ensure the robustness of the observed spectral scaling across different linguistic structures. The text sequences were systematically collected from human-authored sources and correspondingly generated using artificial intelligence.

\subsection{Multilingual Corpus Composition}
For Chinese and English, the dataset covers three primary domains: scientific/technical writing, news reports, and movie reviews; the Chinese corpus additionally includes Gaokao (college entrance exam) style essays.
The human-authored Chinese texts were drawn from CNKI research articles (primarily 2024--2025), CCTV news reports (2022--2025), and Douban movie reviews (e.g., \emph{The Shawshank Redemption}). This subset comprises 1340 samples for BERT-based encoding and 814 for Word2Vec encoding. 
Corresponding AI-generated texts were produced for the same domains, totaling 1062 (BERT) and 761 (Word2Vec) samples.

Human-authored English texts include research papers from journals (e.g., Physical Review Letters/arXiv/Journal of Fluid Mechanics/Physics of Fluids), a cleaned CNN news corpus (2011--2022), and IMDB movie reviews. 
The resulting dataset contains 1228 samples for GPT encoding and 1052 for GloVe encoding.
The AI-generated counterparts consist of 1465 (GPT) and 1263 (GloVe) samples.

The German and Japanese corpora are exclusively restricted to the news domain.
Human German texts ($N = 428$) were sourced from the 10kGNAD news dataset (2015–2016), and the corresponding AI-generated texts contain 428 samples. 
Human Japanese texts ($N=500$) are from a cleaned CC-News subset (2024), with 292 corresponding AI-generated samples.

\subsection{AI Text Generation Protocols}
All AI corpora were generated using the \texttt{DeepSeek-V3} model. 
To ensure that the generated text provided sufficient length for multiscale spectral analysis, prompts were strictly controlled to produce long-form sequences: 600--800+ words per technical paper section, 800+ words per news report (enforcing standard journalistic structures), and over 1,000 words for movie reviews.

\textbf{Scientific Writing (Chinese and English):} 
To cover a diverse semantic phase space, scientific texts were generated across 14 distinct research topics: Generative Artificial Intelligence, Quantum Cryptography, Brain-Computer Interfaces, Edge Computing, Graphene Batteries, 6G Network Architecture, Autonomous Driving Perception Algorithms, Digital Twin Cities, Bioprinting Technology, Zero-Knowledge Proofs, Controlled Nuclear Fusion, Starlink Communication, Exoskeleton Robotics, and Nanomedical Robots. 

The prompt templates enforced strict academic standards, including minimum length, technical specificity, and logical rigor. The Chinese and English prompts were structured as follows, where placeholders like \texttt{\{topic\}} and \texttt{\{section\_title\}} were dynamically injected:

\begin{CJK*}{UTF8}{gbsn}
\begin{quote}
\ttfamily
你正在撰写一篇主题为《\{topic\}》的深度科技论文。
当前请撰写章节：【\{section\_title\}】。

要求：
1. 字数要求：本章节内容必须丰富翔实，字数不少于800字。
2. 内容深度：包含具体的技术参数、理论推导或行业数据，拒绝空泛的套话。
3. 风格：学术性强，逻辑严密。
4. 仅输出本章节的正文内容，不需要重复标题。
\end{quote}
\end{CJK*}

\begin{quote}
\ttfamily
You are writing a comprehensive academic paper on the topic "\{topic\}".
Please write the section: [\{section\_title\}].

 Requirements:
1. Length: The content must be substantial. Please write at least 800 words for this section.
2. Depth: Include specific technical parameters, theoretical derivations, or industry data. Avoid vague generalizations.
3. Style: Professional academic English, rigorous logic.
4. Format: Output ONLY the body text of this section. Do not repeat the title.
\end{quote}

\textbf{Journalistic Reporting (Japanese and German):}
For the news domains, topics were selected to reflect contemporary macroscopic societal and economic themes. The Japanese topics included AI and Future Technology, Global Economy and the Japanese Market, Declining Birthrate and Aging Population, Climate Change, Medical Technology, International Politics, Globalization of the Anime/Manga Industry, Inbound Tourism, Renewable Energy, Educational Reform, Olympic Legacy, and Regional Revitalization. The prompt explicitly required the formal journalistic style (\begin{CJK*}{UTF8}{gbsn}
\emph{常体}
\end{CJK*}) typical of major Japanese publishers (e.g., Asahi, Yomiuri, Nikkei), structured with a lead paragraph, background analysis, and expert commentary:

\begin{CJK*}{UTF8}{min}
\begin{quote}
\ttfamily
テーマ「\{category\}」について、詳細なニュース記事を書いてください。
要件：
1. 具体的な出来事：この分野における架空のニュースイベント、または現在のトレンドを深く分析してください。
2. タイトル：新聞記事らしい、プロフェッショナルで目を引く見出しをつけてください。
3. 構成：リード文（5W1Hを含む）、本文（背景、詳細、データ分析）、専門家のコメント（架空の識者の発言を含む）、結び。
4. 文体：「だ・である」調（常体）を使用し、日本の大手新聞社のスタイルで執筆してください。
5. 長さ：本文は必ず800文字以上にしてください。
\end{quote}
\end{CJK*}

Similarly, the German news generation spanned analogous macroeconomic and societal topics: AI and Industry 4.0, European Markets, Energy Transition, Medical Research, EU Foreign Policy, E-Mobility, Digitalization, Modern Art, Sports Science, Sustainability, Smart Cities, and Food Security. The prompt mandated an objective, sophisticated tone characteristic of major German outlets (e.g., \emph{FAZ, Süddeutsche Zeitung, Der Spiegel}):

\begin{quote}
\ttfamily
Bitte schreiben Sie einen ausführlichen Nachrichtenbericht zum Thema \{category\}.
Anforderungen:
1. Konkretes Ereignis: Erfinden Sie ein realistisches Szenario oder analysieren Sie einen aktuellen Trend.
2. Schlagzeile: Formulieren Sie eine professionelle und prägnante Überschrift.
3. Struktur: Enthalten Sie einen Datumszeile, einen starken Lead-Absatz (W-Fragen), Hauptteil mit Hintergrundinformationen und Analysen, Zitate von Experten (fiktiv aber plausibel) und ein Fazit.
4. Länge: Der Artikel muss mindestens 800 Wörter lang sein. Es soll sich um eine tiefergehende Reportage handeln.
5. Stil: Objektiv, seriös und anspruchsvoll.
\end{quote}

%\nocite{*}

\bibliographystyle{unsrtnat}
\bibliography{B-references}

@string{sam={Stud Appl Math}}

@book{zipf2016human,
  title={Human behavior and the principle of least effort: An introduction to human ecology},
  author={Zipf, George Kingsley},
  year={2016},
  publisher={Ravenio books}
}

@article{nakaishi2024critical,
  title={Critical phase transition in large language models},
  author={Nakaishi, Kai and Nishikawa, Yoshihiko and Hukushima, Koji},
  journal={arXiv preprint arXiv:2406.05335},
  year={2024}
}

@article{lin2017critical,
  title={Critical behavior in physics and probabilistic formal languages},
  author={Lin, Henry W and Tegmark, Max},
  journal={Entropy},
  volume={19},
  number={7},
  pages={299},
  year={2017},
  publisher={MDPI}
}

@article{kolmogorov1991local,
  title={The local structure of turbulence in incompressible viscous fluid for very large {R}eynolds numbers},
  author={Kolmogorov, Andrei Nikolaevich},
  journal={Proceedings of the Royal Society of London. Series A: Mathematical and Physical Sciences},
  volume={434},
  number={1890},
  pages={9--13},
  year={1991},
  publisher={The Royal Society London}
}

@inproceedings{devlin2019bert,
  title={{BERT}: Pre-training of deep bidirectional transformers for language understanding},
  author={Devlin, Jacob and Chang, Ming-Wei and Lee, Kenton and Toutanova, Kristina},
  booktitle={Proceedings of the 2019 conference of the North American chapter of the association for computational linguistics: human language technologies, volume 1 (long and short papers)},
  pages={4171--4186},
  year={2019}
}

@inproceedings{pennington2014glove,
  title={Glo{V}e: Global vectors for word representation},
  author={Pennington, Jeffrey and Socher, Richard and Manning, Christopher D},
  booktitle={Proceedings of the 2014 conference on empirical methods in natural language processing (EMNLP)},
  pages={1532--1543},
  year={2014}
}

@misc{openai2025gptoss120bgptoss20bmodel,
      title={gpt-oss-120b \& gpt-oss-20b Model Card}, 
      author={OpenAI},
      year={2025},
      eprint={2508.10925},
      archivePrefix={arXiv},
      primaryClass={cs.CL},
      url={https://arxiv.org/abs/2508.10925},
}

@misc{word2vecweb,
    author       = {Shen Li},
    title        = {Chinese-Word-Vectors},
    year         = {2023},
    url          = {https://github.com/Embedding/Chinese-Word-Vectors},
    note         = {Accessed: 2025-11-21}
}

@misc{gloveweb,
    author       = {Jeffrey Pennington and Richard Socher and Christopher D. Manning},
    title        = {Glo{V}e: Global Vectors for Word Representation},
    year         = {2024},
    url          = {https://nlp.stanford.edu/projects/glove},
    note         = {Accessed: 2025-11-21}
}

@article{altmann2012origin,
  title        = {On the origin of long-range correlations in texts},
  author       = {Altmann, Eduardo G. and Cristadoro, Giampaolo and Degli Esposti, Mirko},
  journal      = {Proceedings of the National Academy of Sciences},
  year         = {2012},
  volume       = {109},
  number       = {29},
  pages        = {11582--11587},
  doi          = {10.1073/pnas.1117723109}
}

@article{nastrom1985climatology,
  title        = {A climatology of atmospheric wave number spectra of wind and temperature observed by commercial aircraft},
  author       = {Nastrom, Gregory D. and Gage, Kenneth S.},
  journal      = {Journal of the Atmospheric Sciences},
  year         = {1985},
  volume       = {42},
  pages        = {950--960},
  doi          = {10.1175/1520-0469(1985)042<0950:ACOAWS>2.0.CO;2}
}

@article{armstrong1995electron,
  title        = {Electron Density Power Spectrum in the Local Interstellar Medium},
  author       = {Armstrong, John W. and Rickett, Barney J. and Spangler, Steven R.},
  journal      = {The Astrophysical Journal},
  year         = {1995},
  volume       = {443},
  number       = {1},
  pages        = {209--221},
  doi          = {10.1086/175515}
}

@article{goldstein1995mhd,
  title        = {Magnetohydrodynamic Turbulence in the Solar Wind},
  author       = {Goldstein, Melvyn L. and Roberts, David A. and Matthaeus, William H.},
  journal      = {Annual Review of Astronomy and Astrophysics},
  year         = {1995},
  volume       = {33},
  pages        = {283--325},
  doi          = {10.1146/annurev.aa.33.090195.001435}
}

@article{vaswani2017attention,
  title={Attention is all you need},
  author={Vaswani, Ashish and Shazeer, Noam and Parmar, Niki and Uszkoreit, Jakob and Jones, Llion and Gomez, Aidan N and Kaiser, {\L}ukasz and Polosukhin, Illia},
  journal={Advances in neural information processing systems},
  volume={30},
  year={2017}
}

@inproceedings{brown2020gpt3,
  title        = {Language Models are Few-Shot Learners},
  author       = {Brown, Tom B. and Mann, Benjamin and Ryder, Nick and Subbiah, Melanie and Kaplan, Jared and Dhariwal, Prafulla and Neelakantan, Arvind and Shyam, Pranav and Sastry, Girish and Askell, Amanda and Agarwal, Sandhini and Herbert-Voss, Ariel and Krueger, Gretchen and Henighan, Tom and Child, Rewon and Ramesh, Aditya and Ziegler, Daniel M. and Wu, Jeffrey and Winter, Clemens and Hesse, Christopher and Chen, Mark and Sigler, Eric and Litwin, Mateusz and Gray, Scott and Chess, Benjamin and Clark, Jack and Berner, Christopher and McCandlish, Sam and Radford, Alec and Sutskever, Ilya and Amodei, Dario},
  booktitle    = {Advances in Neural Information Processing Systems (NeurIPS)},
  year         = {2020},
  volume       = {33},
  pages        = {1877--1901},
  url          = {https://proceedings.neurips.cc/paper/2020/hash/1457c0d6bfcb4967418bfb8ac142f64a-Abstract.html}
}

@inproceedings{mikolov2013distributed,
  title        = {Distributed Representations of Words and Phrases and their Compositionality},
  author       = {Mikolov, Tomas and Sutskever, Ilya and Chen, Kai and Corrado, Greg and Dean, Jeffrey},
  booktitle    = {Advances in Neural Information Processing Systems (NeurIPS)},
  year         = {2013},
  volume       = {26},
  pages        = {3111--3119},
  url          = {https://papers.nips.cc/paper/5021-distributed-representations-of-words-and-phrases-and-their-compositionality}
}

\end{document}